\documentclass[final]{cvpr}
\usepackage{times}
\usepackage{epsfig}
\usepackage{graphicx}
\usepackage{amsmath}
\usepackage{amssymb}
\usepackage{multirow}
\usepackage{booktabs} 
\usepackage{paralist}	
\usepackage{xcolor}
\usepackage{enumerate}

\usepackage[pagebackref=true,breaklinks=true,colorlinks,bookmarks=false]{hyperref}

\def\eg{\emph{e.g.}} 

\def\ie{\emph{i.e.}}

\def\etal{\emph{et~al.}}
\definecolor{almond}{rgb}{0.94, 0.6, 0.4}
\def\bmatrix#1{\left[ \begin{matrix} #1 \end{matrix} \right]}  
\newcommand{\T}{\top}  
\newcommand\blfootnote[1]{%
  \begingroup
  \renewcommand\thefootnote{}\footnote{#1}%
  \addtocounter{footnote}{-1}%
  \endgroup
}

\def\cvprPaperID{3297} 
\def\confYear{CVPR 2021}

\begin{document}

\def\YD#1{{\color{red}{\bf [Yuchao:} {{#1}}{\bf ]}}}
\def\JY#1{{\color{blue}{\bf [JY:} {{#1}}{\bf ]}}}
\def\HL#1{{\color{green}{\bf [HL:} {{#1}}{\bf ]}}}
\def\YR#1{{\color{almond}{\bf [YR:} {{#1}}{\bf ]}}}


\title{Deep Two-View Structure-from-Motion Revisited}


\author{\textsuperscript{*}Jianyuan Wang$^{1}$, \textsuperscript{*}Yiran Zhong$^{1}$, Yuchao Dai$^{2}$, Stan Birchfield$^{3}$, \\
Kaihao Zhang$^{1}$, Nikolai Smolyanskiy$^{3}$, Hongdong Li$^{1}$ \\
$^{1}$Australian National University, 
$^{2}$Northwestern Polytechnical University,
$^{3}$NVIDIA}

\maketitle
\pagestyle{empty}  
\thispagestyle{empty} 

\blfootnote{* indicates equal contribution, listed in alphabetical order. Yiran is the corresponding author. Work was partially done when Yiran was an intern at NVIDIA, Redmond, WA.}


\begin{abstract}

Two-view structure-from-motion (SfM) is the cornerstone of 3D reconstruction and visual SLAM.
Existing deep learning-based approaches formulate the problem by either recovering absolute pose scales from two consecutive frames or predicting a depth map from a single image, both of which are ill-posed problems. 
In contrast, we propose to revisit the problem of deep two-view SfM by leveraging the well-posedness of the classic pipeline.
Our method consists of 1) an optical flow estimation network that predicts dense correspondences between two frames; 2) a normalized pose estimation module that computes relative camera poses from the 2D optical flow correspondences, and 3) a scale-invariant depth estimation network that leverages epipolar geometry to reduce the search space, refine the dense correspondences, and estimate relative depth maps.
Extensive experiments show that our method outperforms all state-of-the-art two-view SfM methods by a clear margin on KITTI depth, KITTI VO, MVS, Scenes11, and SUN3D datasets in both relative pose and depth estimation.

\end{abstract}

\vspace{-3mm}

\section{Introduction}

\begin{figure*}[t]
\begin{center}
  \includegraphics[width=\linewidth]{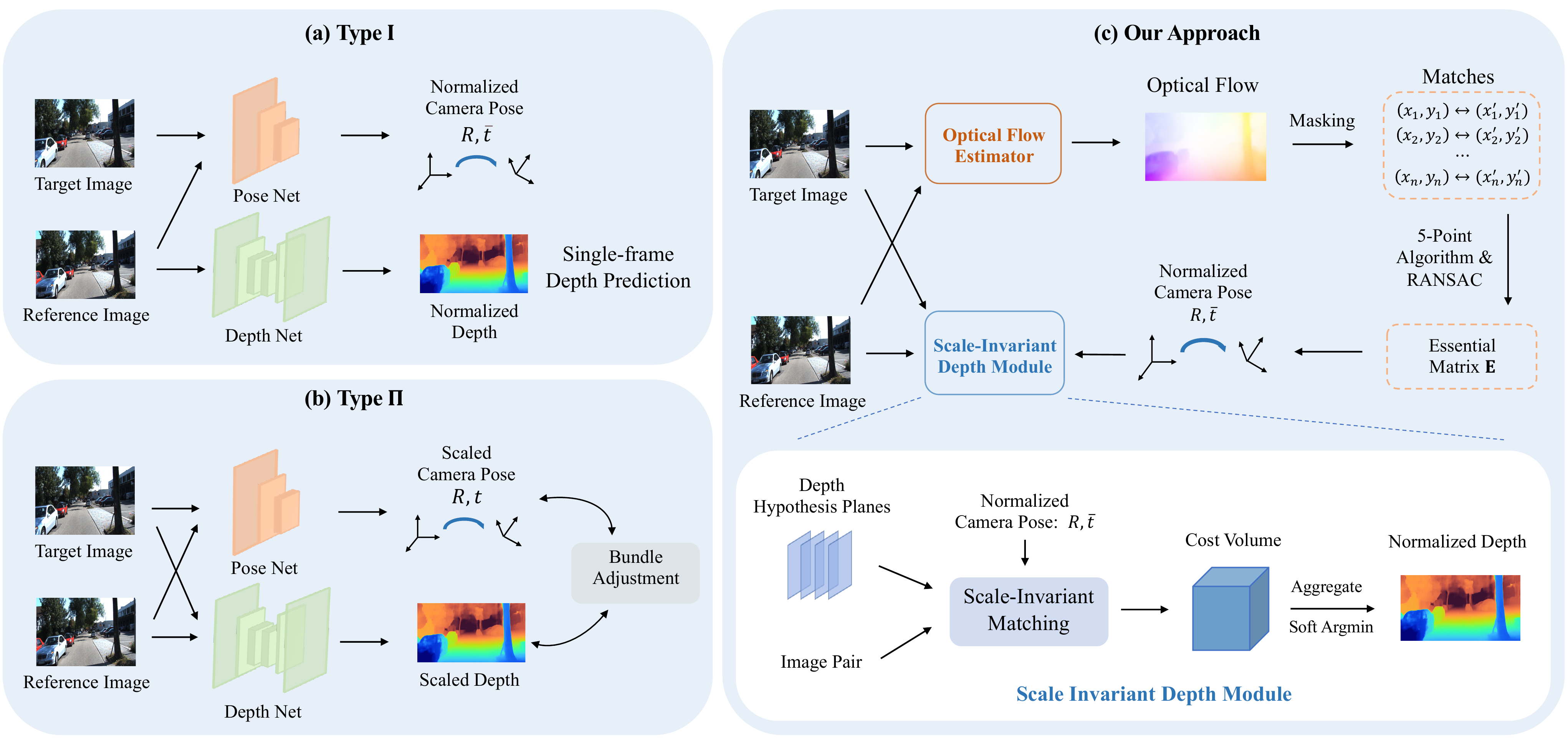}
\end{center}
\vspace{-4mm}
  \caption{\small \textbf{Comparison between our method and previous deep monocular structure-from-motion methods.} We formulate camera pose estimation as a 2D matching problem (optical flow) and depth prediction as a 1D matching problem along an epipolar line. In contrast, previous methods suffer from ill-posedness (either single-frame depth prediction, in the case of Type $\mathbf{I}$, or scaled estimates, in the case of Type $\mathbf{II}$).}
\label{fig:compare_method_type}
\vspace{-4mm}
\end{figure*}

Two-view structure-from-motion (SfM) is the problem of estimating the camera motion and scene geometry from two image frames of a monocular sequence.
As the foundation of both 3D reconstruction and visual simultaneous localization and mapping (vSLAM), this important problem finds its way into a wide range of applications, including autonomous driving, augmented/virtual reality, and robotics.

Classic approaches to two-view SfM follow a standard pipeline of first matching features/edges between the two images, then inferring motion and geometry from those matches~\cite{davison2007pami:monoslam,klein2007ismar:ptam,mur2015orbslam,forster2014icra:svo,engel14eccv:lsdslam,engel2016dso,zhong2017self,zhong2020displacement}.
When imaging conditions are well-behaved (constant lighting, diffuse and rigid surfaces, and non-repeating visual texture), the matching process is well-posed.
And, once the matches have been found, the motion and geometry can be recovered.

For decades, researchers who work in this area have generally required at least two views, and their methods have recovered only \emph{relative} camera motion and \emph{relative} scene geometry (that is, shape up to an unknown scale factor).
Without a \emph{priori} knowledge of scale or recognizable objects in the scene, it is impossible to recover scene geometry from a single view~\cite{hassner2006cvprw,zhong18eccvmono,xie2019iccv:pix2vox,li2019bmvc:svosr}.  Similarly, it is impossible to infer \emph{absolute} scale from two views of a scene~\cite{hartley2003multiple}.

With the rise of deep learning, a number of researchers have recently explored neural network-based solutions to two-view SfM.
Most of these methods fall into one of two categories.
In the first category, which we shall call Type $\mathbf{I}$, the problem is treated as a joint optimization task of monocular depth and pose regression~\cite{sfmlearner,yin2018geonet,Ranjan_2019_CVPR}.
Two networks are used:  one to estimate the \emph{up-to-scale} depth from a single image, and the other one to predict the \emph{up-to-scale} camera pose from two input images. Both networks act independently during inference.
In the second category, denoted Type $\mathbf{II}$, the \emph{scaled} camera pose and the \emph{scaled} depth are inferred from the image pair, and are iteratively refined via multi-view geometry~\cite{ummenhofer2017demon,tang2018ba,teed2018deepv2d}. 
While the power of deep learning allows both Type $\mathbf{I}$ and Type $\mathbf{II}$ solutions to achieve compelling results,
we note that their formulations attempt to solve one of the \emph{ill-posed}~\cite{bertero1988illposed} problems mentioned above.


In this paper, we revisit the use of deep learning for two-view SfM.
Our framework follows the classic SfM pipeline that features are matched between image frames to yield \emph{relative} camera poses, from which \emph{relative} depths are then estimated. 
By combining the strengths of deep learning within a classic pipeline, we are able to avoid ill-posedness, which allows our approach to achieve state-of-the-art results on several benchmarks.

A comparison between our approach and existing pipelines is shown in Fig.~\ref{fig:compare_method_type}.
Our method operates by first estimating dense matching points between two frames using a deep optical flow network~\cite{zhong2019unsupervised,zhong2020nipsflow}, from which a set of highly reliable matches are sampled in order to compute the relative camera pose via a GPU-accelerated classic five-point algorithm~\cite{li2006five} with RANSAC~\cite{fischler1981random}. Since these relative camera poses have scale ambiguity, the estimated depth suffers from scale ambiguity as well. Therefore, in order to supervise the estimated scale-ambiguous depth with the (scaled) ground truth depth, we propose a scale-invariant depth estimation network combined with scale-specific losses to estimate the final relative depth maps. Since the search space of the depth estimation network is reduced to epipolar lines thanks to the camera poses, it yields higher accuracy than directly triangulating the optical flows with the estimated camera poses. 
We demonstrate the effectiveness of our framework by achieving state-of-the-art accuracy in both pose and depth estimation on KITTI depth, KITTI VO, MVS, Scenes11, and SUN3D datasets. 

Our main contributions are summarized as:
\begin{compactenum}[1)]
    \item We revisit the use of deep learning in SfM, and propose a new deep two-view SfM framework that avoids ill-posedness. Our framework combines the best of deep learning and classical geometry.
    \item We propose a scale-invariant depth estimation module to handle the mismatched scales between ground truth depth and the estimated depth.
    \item Our method outperforms all previous methods on various benchmarks for both relative pose estimation and depth estimation under the two-view SfM setting.
\end{compactenum}

\section{Two-view Geometry: Review and Analysis}

The task of two-view SfM refers to estimating the relative camera poses and dense depth maps from two consecutive monocular frames. In classic geometric vision, it is well understood that the camera poses as well as the depth maps can be computed from image matching points alone without any other information \cite{longuet1981computer}.\footnote{Excluding degenerate cases.} 
 
Given a set of image matching points in homogeneous coordinates, $\mathbf{x}_i = \bmatrix{x_i & y_i & 1}^\T$ and $\mathbf{x}^\prime_i = \bmatrix{x'_i & y'_i & 1}^\T$ with known camera intrinsic matrix $\mathbf{K}$, the two-view SfM task is to find a camera rotation matrix $\mathbf{R}$ and a translation vector $\mathbf{t}$ as well as the corresponding 3D homogeneous point $\mathbf{X}_i$ such that:
\begin{equation}
    \mathbf{x}_i = \mathbf{K}\bmatrix{\mathbf{I}\,\,|\,\,\mathbf{0}}\mathbf{X}_i \quad \mathbf{x}_i^\prime = \mathbf{K}\bmatrix{\mathbf{R}\,\,|\,\,\mathbf{t}}\mathbf{X}_i  \qquad \forall i.
\end{equation}

A classical method to solve this problem consists of three consecutive steps: 1) Computing the essential matrix $\mathbf{E}$ from the image matching points $\mathbf{x}_i$ and $\mathbf{x}^\prime_i$; 2) Extracting the relative camera pose $\mathbf{R}$ and $\mathbf{t}$ from the essential matrix $\mathbf{E}$; 3) Triangulating the matching points $\mathbf{x}_i$ and $\mathbf{x}^\prime_i$ with the camera pose to get the 3D point $\mathbf{X}_i$.   

All steps in this pipeline are \emph{well-posed} problems. The essential matrix $\mathbf{E}$ can be solved with at least 5 matching points using the equation below:
\begin{equation}
\mathbf{x}^{\prime\T}_i\mathbf{K}^{-\T}\mathbf{E}\mathbf{K}^{-1}\mathbf{x}_i=0 \qquad \forall i.
\label{eq:fmatrix}
\end{equation}
$\mathbf{R}$ and $\mathbf{t}$ can be computed from $\mathbf{E}$ using matrix decomposition such that $\mathbf{E} = \mathbf{SR}$, where $\mathbf{S}$ is a skew symmetric matrix and $\mathbf{R}$ is a rotation matrix. Since for any non-zero scaling factor $\alpha$, $\bmatrix{\alpha\mathbf{t}}_{\times} \mathbf{R} =  \alpha\bmatrix{\mathbf{t}}_{\times} \mathbf{R} = \alpha\mathbf{E}$ provides a valid solution, there is a scale ambiguity for relative camera pose estimation. The 3D point $\mathbf{X}_i$ can be computed by triangulation with a global scale ambiguity.

The method above assumes the ideal case in which all image points are perfectly matched. To handle mismatched points in real scenarios, researchers have established a classical standard pipeline to estimate geometry information from two consecutive frames \cite{hartley2003multiple}.

\subsection{The Classic Standard Pipeline}
With decades of development and refinement, the classic standard pipeline~\cite{hartley2003multiple} is widely used in many conventional state-of-the-art SfM and vSLAM systems~\cite{mur2015orbslam,colmap,agarwal2011building}. Since almost all geometry information can be recovered from image matching points, the key is to recover a set of (sparse or dense) accurate matching points. To this end, the pipeline often starts with sparse (or semi-dense) distinct feature extraction and matching to get sparse matching points, as sparse matching is more accurate than dense matching. To further refine the matching results, the RANSAC scheme~\cite{fischler1981random} is used to filter the matching points that do not fit the majority motion. These outliers often include mismatches and dynamic objects in a scene. After retrieving the camera poses from the refined matching points, the depth of these points can be computed via triangulation. In some cases, if it is desired to estimate dense depth maps rather than the sparse 3D points, multi-view stereo matching algorithms can be used to recover the dense depth maps with the estimated camera poses.

The Achilles' heel of this pipeline is therefore the matching of points. Conventional matching algorithms often suffer from low accuracy on non-Lambertian, blurry, and textureless surfaces. However, this shortage can be largely alleviated by deep learning~\cite{sun2018pwc,teed2020raft,zhong2020nipsflow, Zhong_2018_ECCV,cheng2020hierarchical,li2021arvo}. With sufficient training data, such networks can learn to handle these scenarios. 
In our proposed approach, we leverage a deep optical flow network~\cite{zhong2020nipsflow} to compute these correspondences.

\subsection{Deep Learning based Methods}

As discussed earlier, two-view SfM requires to estimate both camera poses and dense depth maps.  Existing deep learning based methods either formulate the problem as pose and monocular depth regression (Type~$\mathbf{I}$) or as pose regression and multi-view stereo matching (Type~$\mathbf{II}$). We analyze both types of methods below.

\textbf{Type~$\mathbf{I}$ methods}
consist of a monocular depth estimation network and a pose regression network. The two-view geometry constraints are used as self-supervisory signals to regularize both camera poses and depth maps~\cite{sfmlearner,yin2018geonet,Mahjourian_2018_CVPR,chen2019self,GANVO,Ranjan_2019_CVPR}. 
As a result, most of these approaches are self-supervised.
Because single-view depth estimation is inherently ill-posed, as discussed earlier, these methods are fundamentally limited by how well they can solve that challenging problem. They rely on the priors from the training data to predict depth only given a single image.

Moreover, since the two-view geometry constraints are only suitable for a stationary scene, SfMLearner~\cite{sfmlearner} simultaneously estimates an explainability mask to exclude the dynamic objects while GeoNet~\cite{yin2018geonet} utilizes an optical flow module to mask out these outliers by comparing the rigid flow (computed by camera poses and depth maps) with the non-rigid flow (computed by the optical flow module). Other methods focus on implementing more robust loss functions, such as ICP loss~\cite{Mahjourian_2018_CVPR}, motion segmentation loss~\cite{Ranjan_2019_CVPR}, or epipolar loss~\cite{chen2019self}. 

\textbf{Type~$\mathbf{II}$ methods}
require two image frames to estimate depth maps and camera poses at test time (unlike Type I methods, which estimate depth from a single frame).
Most supervised deep methods fall into this category. 
As a pioneer of this type, DeMoN~\cite{ummenhofer2017demon} concatenates a pair of frames and uses multiple stacked encoder-decoder networks to regress camera poses and depth maps, implicitly utilizing multi-view geometry. 

Similar strategies have been adapted by \cite{tang2018ba,clark2018ls,flowmotion,teed2018deepv2d} through replacing generic layers between camera poses and depth maps with optimization layers that explicitly enforce multi-view geometry constraints. For example, BANet~\cite{tang2018ba} parameterizes dense depth maps with a set of depth bases~\cite{zhong2020efficient} and imposes bundle adjustment as a differentiable layer into the network architecture. Wang~\etal~\cite{flowmotion} use regressed camera poses to constrain the search space of optical flow, estimating dense depth maps via triangulation.  
DeepV2D~\cite{teed2018deepv2d} separates the camera pose and depth estimation, iteratively updating them by minimizing geometric reprojection errors. Similarly, DeepSFM~\cite{wei2019deepsfm} initiates its pose estimation from DeMoN~\cite{ummenhofer2017demon}, sampling nearby pose hypotheses to bundle adjust both poses and depth estimation. Nevertheless, with the ground truth depth as supervision, it requires the pose regression module to estimate camera poses with absolute scale, which is generally impossible from a pair or a sequence of monocular frames alone~\cite{hartley2003multiple}. To 
mitigate this ill-posed problem, they utilize dataset priors and semantic knowledge of the scene to estimate the absolute scale.

\section{Method}
\label{sec_method}
In this section, we propose a new deep two-view SfM framework that aims to address the Achilles' heel of the classical SfM pipeline (\emph{viz.}, matching) via deep learning. 
Our method is able to find better matching points and therefore more accurate poses and depth maps, especially for textureless and occluded areas. 
At the same time, it follows the wisdom of classic methods to avoid the ill-posed problems.
By combining the best of both worlds, our approach is able to achieve state-of-the-art results, outperforming all previous methods by a clear margin.

Following the classic standard pipeline~\cite{hartley2003multiple}, we formulate the two-frame structure-from-motion problem as a three-step process:  1) match corresponding points between the frames, 2) estimate the essential matrix and hence the relative camera pose, and 3) estimate dense depth maps up to an unknown scale factor.
These steps, along with the loss function used for training, are described in more detail in the following subsections.

\subsection{Optical Flow Estimation}

As a fundamental problem in computer vision, optical flow estimation has been extensively studied for several decades~\cite{horn1981determining}. 
With the recent progress in deep learning, deep optical flow methods now dominate various benchmarks~\cite{kittidataset,Butler2012Sintel} and can handle large displacements as well as textureless, occluded, and non-Lambertian surfaces. In our framework, we utilize the state-of-the-art network, DICL-Flow~\cite{zhong2020nipsflow}, to generate dense matching points between two consecutive frames. 
This method uses a displacement-invariant matching cost learning strategy and a soft-argmin projection layer to ensure that the network learns dense matching points rather than image-flow regression.
The network was trained on synthetic datasets (FlyingChairs~\cite{flyingchair} and FlyingThings~\cite{Mayer2016Things3D}) to avoid data leakage, \ie, the network was not trained on any of the test datasets.

\subsection{Essential Matrix Estimation}

The traditional approach to estimating camera pose between two image frames is to match sparse points, \eg, SIFT features~\cite{loweSIFT}. 
Then, given a set of matching points $\mathbf{x}\leftrightarrow \mathbf{x}^\prime$ and the camera intrinsic matrix $\mathbf{K}$, the essential matrix~\cite{longuet1981computer} $\mathbf{E}$ can be recovered from the five-point algorithm~\cite{nister2004efficient,li2006five}. 
By decomposing the essential matrix as $\mathbf{E} = [\mathbf{t}]_\times\mathbf{R}$, the rotation matrix $\mathbf{R}$ and the translation vector $\mathbf{t}$ can be recovered up to a scale ambiguity. 
Conventionally, outliers in the matching points are filtered using robust fitting techniques such as RANSAC~\cite{fischler1981random}. RANSAC repeatedly estimates the essential matrix from randomly sampled minimal matching sets and selects the solution that is satisfied by the largest proportion of matching points under a certain criterion.

Unlike all previous deep learning-based methods that regress the camera poses from input images, we use matching points to compute the camera poses. 
The key question is this:  How to robustly filter the noisy dense matches from optical flow in order to retain only the high quality matches?  
There are multiple ways to filter out unreliable matching points such as flow uncertainty, consistency check, or using a network to regress a mask. 
Empirically, we find that simply using SIFT keypoint locations (note that we do \emph{not} use SIFT matching) to generate a mask works well in all datasets. 
The hypothesis is that optical flow is more accurate in rich textured areas. 
The optical flow matches at the locations within the mask
are filtered by RANSAC with GPU acceleration, to avoid distraction by dynamic objects.
After retrieving the essential matrix $\mathbf{E}$, the camera pose ($\mathbf{R},\mathbf{t}$) is recovered using matrix decomposition.

\subsection{Scale-Invariant Depth Estimation}

Once we have recovered the up-to-scale relative camera pose, with the dense matching points from optical flow estimation, we could compute the dense depth map by performing triangulation. 
However, such an approach would not take advantage of the epipolar constraint.
As a result, we perform the matching again by constraining the search space to epipolar lines computed from the relative camera poses. 
This process is similar to multi-view stereo (MVS) matching with one important difference: we do not have the absolute scale in inference. 
With the up-to-scale relative pose, if we were to directly supervise the depth estimation network with ground truth depth, there would be a mismatch between the scale of the camera motion and the scale of the depth map. 

\textbf{Previous approaches.}  To resolve this paradox, previous methods either use a scale-invariant loss~\cite{eigen2014depth} or regress the absolute scale with a deep network~\cite{ummenhofer2017demon,teed2018deepv2d,wei2019deepsfm}. 

The scale-invariant loss $\ell_{\text{SI}}$ is defined as:
\begin{equation}
\ell_{\text{SI}} = \sum_{\mathbf{x}} \left(\text{log}(d_\mathbf{x}) - \text{log}(\hat{d}_\mathbf{x}) + \eta(d, \hat{d}) \right)^2,
\end{equation}
where $d_\mathbf{x}$ and $\hat{d}_\mathbf{x}$ are the ground truth and estimated depth, respectively, at pixel $\mathbf{x}$; and \scalebox{0.9}{$\eta(d, \hat{d}) = \frac{1}{N} \sum_{\mathbf{x}} \left(\text{log}(\hat{d}_\mathbf{x}) - \text{log}(d_\mathbf{x}) \right)$}, where $N$ is the number of pixels, measures the mean log difference between the two depth maps. While working for the direct depth regression pipelines, the scale-invariant loss introduces an ambiguity for network learning as the network could output depth maps with different scales for each sample. This loss may hinder the principle of plane-sweep, where depth maps with consistent scale across frames of the sequence are desired.
Plane sweep~\cite{im2019dpsnet} is the process that enforces the epipolar constraint, which reduces the search space from 2D to 1D. 

Plane-sweep powered networks require consistent scale during the training and testing process. For example, if we train a network with absolute scales and test it with a normalized scale, its performance will drop significantly (we provide an ablation study in Section~\ref{sec:ablaiton}). Since it is impossible to recover the absolute scale from two images, some previous methods~\cite{ummenhofer2017demon,teed2018deepv2d,wei2019deepsfm} use a network to regress a scale to mimic the absolute scale in inference. This strategy slightly alleviates the scale paradox at the cost of making the problem ill-posed again.

\textbf{Scale-Invariant Matching.}  To solve this paradox and keep the problem well-posed, we propose a scale-invariant matching process to recover the up-to-scale dense depth map.
Mathematically, given an image point $\mathbf{x}$, we generate $L$ matching candidates $\{\mathbf{x}^\prime_{l}\}_{l=1}^L$:
\begin{equation}
    \mathbf{x}^\prime_{l}\sim\mathbf{K}[\mathbf{R}\,\,|\,\,\mathbf{t}]\bmatrix{
(\mathbf{K}^{-1}\mathbf{x})d_l\\
1},
\label{eq:warping}
\end{equation}
where $d_l = (L\times d_{\text{min}})/l,~(l=1,...,L)$ is the depth hypothesis and $d_{\text{min}}$ is a fixed minimum depth. 
\begin{figure}[t]
\begin{center}
  \includegraphics[width=\linewidth]{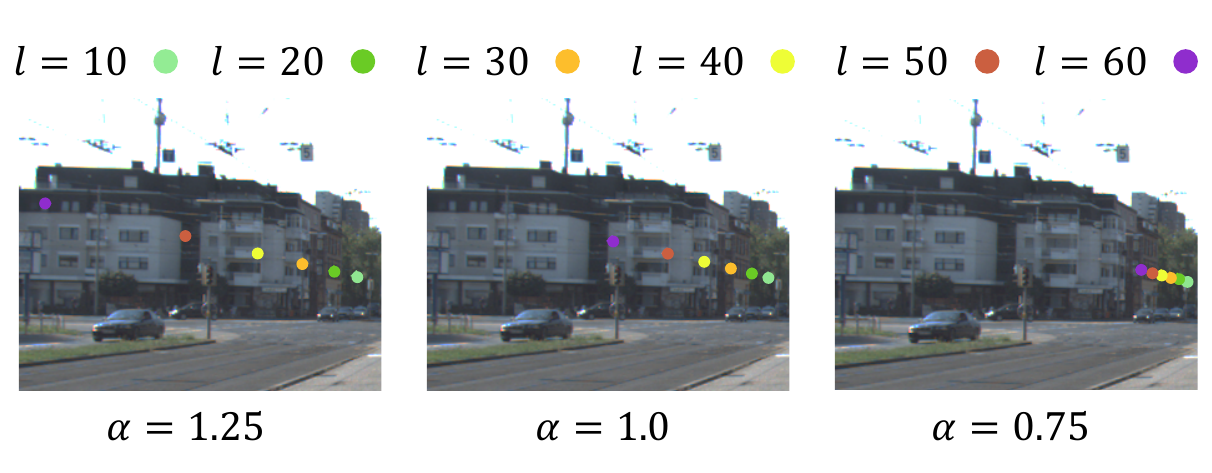}
\end{center}
\vspace{-4mm}
  \caption{\small \textbf{The Effect of Various Scale Factors during Plane Sweep.} For a certain pixel, we visualize its six depth hypotheses with different colors in the target frame. As the scale factor $\alpha$ changes, the sampling distribution varies.}
\label{fig:scale_plane_sweep}
\vspace{-4mm}
\end{figure}

In the standard plane-sweep setting, the sampling distribution of matching candidates varies depending on the scale factor $\alpha = \left\|\mathbf{t}\right\|_2 $, as illustrated in Fig.~\ref{fig:scale_plane_sweep}. Instead, we normalize the translation vectors to $\mathbf{\overline{t}} \equiv \mathbf{t}/\alpha$ such that $\left\|\mathbf{\overline{t}}\right\|_2 = 1$, since we do not know the absolute scale in our problem. Substituting the normalized translation $\mathbf{\overline{t}}$ for $\mathbf{t}$ in Eq.~\eqref{eq:warping}, with fixed $\{{d}_{l}\}_{l=1}^L$, the distribution of matching candidates $\{\mathbf{x}^\prime_{l}\}_{l=1}^L$ are now invariant to scale.

To make the estimated and ground truth depths compatible, according to Eq.~\eqref{eq:warping}, we need to scale the estimated depth $\hat{\mathbf{d}}$ correspondingly to match the ground truth depth $\mathbf{d}$: 
\begin{equation}
    \mathbf{d}  \sim {\alpha_{\text{gt}}}\hat{\mathbf{d}}, 
\end{equation}
where $\alpha_{\text{gt}}$ refers to the ground truth scale.

This scale-invariant matching strategy plays a crucial role in our framework as it makes our network no longer suffer from the scale misalignment problem. Please note our competitors cannot benefit from scale-invariant matching because they usually avoid the scale misalignment problem by predicting absolute scales. A detailed discussion is provided in Section~\ref{sec:ablaiton}.

\begin{table*}[tp]
\small
\tabcolsep=0.23cm
\centering
\caption{\small{\textbf{Depth Evaluation on KITTI Depth Dataset.} We compare our results to state-of-the-art single-frame depth estimation methods and deep SfM methods on the KITTI depth Eigen split. We evaluate all SfM methods under two-view SfM setting for a fair comparison. The ``Eigen SfM'' split ($256$ frames) excludes frames that are close to static or contain many dynamic objects in the Eigen split. The type \textbf{S} means supervised single frame depth estimation. Note that Type \textbf{I} methods are self-supervised methods. Bold indicates the best.
}} 
\begin{tabular}{c| c | c || c | c | c |c| c | c | c | c }
\hline
\multirow{ 2 }{*}{Split}& \multirow{ 2 }{*}{Type} & \multirow{ 2 }{*}{Method}  &  \multicolumn{5}{c}{ lower is better } & \multicolumn{3}{ |c  }{ higher is better }	  \\ 
\cline{4-11}
& & & Abs Rel & Sq Rel & $\text{RMSE}$ & $\text{RMSE}_{log}$ & $\text{D1-all}$ & $\delta < 1.25$ & $\delta < 1.25^{2}$ & $\delta < 1.25^{3}$   \\
\hline\hline
\multirow{11}{*}{\rotatebox{90}{{Eigen}}} &\multirow{2}{*}{\textbf{S}}  & 
DORN \cite{fu2018deep} & {0.072} &  {0.307} & {2.727} & {0.120} &0.163& {0.932} & {0.984} & {0.994}  \\ 
& & VNL \cite{yin2019enforcing} & 0.072&-&3.258&0.117&0.176& 0.938 &\textbf{0.990}&\textbf{0.998}\\
\cline{2-11}
&\multirow{5}{*}{\textbf{I}} & SfMLearner~\cite{sfmlearner} &0.208 &1.768 &6.856 &0.283 & - & 0.678 &0.885 &0.957\\
& &GeoNet~\cite{yin2018geonet} &0.155 &1.296 &5.857 &0.233 &- &0.793 &0.931 &0.973\\
& &CCNet~\cite{Ranjan_2019_CVPR} &0.140 &1.070 &5.326 &0.217 &- &0.826 &0.941 &0.975\\
& &GLNet~\cite{chen2019self} &0.099 &0.796 &4.743 &0.186 &- &0.884 &0.955 &0.979\\
\cline{2-11}
&\multirow{2}{*}{{{\textbf{II}}}} & BANet \cite{tang2018ba} & {0.083} &  - & {3.640} & {0.134} &-& {-} & {-} & {-}  \\ 
& & DeepV2D \cite{teed2018deepv2d} & 0.064 &0.350 &2.946&0.120&0.142&0.946&0.982&0.991 \\ \cline{2-11}
& & Ours & \textbf{0.055} &\textbf{0.224} &\textbf{2.273}&\textbf{0.091}&\textbf{0.107}&\textbf{0.956}&0.984&0.993 \\
\hline\hline
\multirow{4}{*}{\rotatebox{90}{{Eigen SfM}}} &\multirow{2}{*}{\textbf{S}} & DORN \cite{fu2018deep}  & 0.067&0.295& 2.929& 0.108 & 0.130 &0.949 &0.988 & 0.995 \\ 
& & VNL \cite{yin2019enforcing}  &0.065&0.297& 3.172 &0.106&0.168& 0.945 &0.989 & 0.997 \\ 
\cline{2-11}
&\textbf{II} & DeepV2D\cite{teed2018deepv2d}  & 0.050 & 0.212& 2.483 & 0.089 & 0.091 & 0.973& 0.992 & 0.997 \\ \cline{2-11}
& & Ours  & \textbf{0.034} &\textbf{0.103}& \textbf{1.919} & \textbf{0.057} &\textbf{0.031}& \textbf{0.989}& \textbf{0.998}& \textbf{0.999}\\
 \hline
\end{tabular}
\label{tab:kitti_depth}
\vspace{-2mm}
\end{table*}


\subsection{Loss Function}

Our framework is trained in an end-to-end manner with the supervision of ground truth depth maps and ground truth scales. Given a predicted depth $\hat{\mathbf{d}}$ and a ground truth depth $\mathbf{d}$, we supervise the depth using the Huber loss:
\begin{equation}
    \mathcal{L}_{\text{depth}}  = \sum_\mathbf{x} \ \ell_{\text{huber}}\left(\alpha_{\text{gt}} \hat{\mathbf{d}}_\mathbf{x} - \mathbf{d}_\mathbf{x}\right),
\end{equation}
where $\ell_{\text{huber}}(z) = 0.5z^2 \text{ if } |z| <1, |z-0.5| \text{ otherwise}$.
It should be noted that our predicted depth is up-to-scale and does not require ground truth scale at inference time.

If both ground truth camera pose $(\mathbf{R}, \mathbf{t})$ and ground truth depth $d_{\mathbf{x}}$ are given, we can also update the optical flow network by computing the rigid flow $\mathbf{u}_\mathbf{x} \equiv \mathbf{x}^{\prime} - \mathbf{x}$ for 2D point $\mathbf{x}$:
\begin{equation}
\mathbf{x}^{\prime} \sim \mathbf{K}[\mathbf{R}\,\,|\,\,\mathbf{t}]\bmatrix{(\mathbf{K}^{-1}\mathbf{x})d_{\mathbf{x}}\\1}.
    \label{eq:flow_sup}
\end{equation}

The rigid flow can work as a supervision signal, computing the $\ell_2$ distance with the estimated optical flow $\hat{\mathbf{u}}_\mathbf{x}$:

\begin{equation}
    \mathcal{L}_{\text{flow}}  = \sum_\mathbf{x} \ \left(\hat{\mathbf{u}}_\mathbf{x} - \mathbf{u}_\mathbf{x}\right)^2.
    \label{eq:flow_loss}
\end{equation}
The total loss function of our framework is then given by:
\begin{equation}
   \mathcal{L}_{\text{total}} = \mathcal{L}_{\text{depth}} + \lambda \mathcal{L}_{\text{flow}}.
\end{equation}
We set $\lambda = 1$ to fine-tune the optical flow estimator, or $\lambda = 0$ to use the flow model pretrained on synthetic datasets.


\section{Experiments}

In this section, we provide quantitative and qualitative results of our framework on various datasets, showing comparison with state-of-the-art SfM methods. We also provide an extensive ablation study to justify our framework design. Due to the scale-ambiguity nature of the two-view SfM problem, we scale the results of ours and others using the same scaling strategy as in~\cite{teed2018deepv2d,tang2018ba}. 
For all experiments, our optical flow estimator is \cite{zhong2020nipsflow} while the architecture of depth estimator is based on \cite{im2019dpsnet}. Implementation details (such as hyperparameters of the optimizer or network) are provided in the supplementary material.

\subsection{Datasets}

\textbf{KITTI Depth}~\cite{kittidataset} is primarily designed for monocular depth evaluation in autonomous driving scenarios, which does not take camera motions and dynamic objects into account. The Eigen split~\cite{eigen2014depth}, which contains $697$ single frames for testing, is a widely used split for evaluating monocular depth estimation. To adapt it for two-view SfM evaluation, we pair nearby frames. Also, since the Eigen split contains a number of frames with nearly static camera motion or many moving objects (which lead to ill-posed situations for two-view SfM), we filter out these frames to create an Eigen SfM split ($256$ frames) to evaluate SfM algorithms in  well-conditioned scenarios. Specifically, we first pair each frame with its next frame then manually remove these pairs with small relative translations (less than $0.5$ meters) or contain large dynamic objects\footnote{We define a dynamic object which occupies more than $20\%$ pixels of a scene as a large dynamic object.}.


\textbf{KITTI VO}~\cite{kittidataset} is primarily used for evaluating camera pose estimation. It contains ten sequences (more than 20k frames) with ground truth camera poses. According to the setting of \cite{sfmlearner}, we test our pose estimation accuracy on all 2700 frames of the ``$09$'' and ``$10$'' sequences, using consecutive frames from the left camera.

\textbf{MVS, Scenes11, and SUN3D.} MVS is collected from several outdoor datasets by~\cite{ummenhofer2017demon}. Different from KITTI which is built through video sequences with close scenes, MVS has outdoor scenes from various sources.
{Scenes11}~\cite{ummenhofer2017demon} is a synthetic dataset generated by random shapes and motions. It is therefore annotated with perfect depth and pose, though the images are not realistic.
{SUN3D}~\cite{xiao2013sun3d} provides indoor images with noisy depth and pose annotation. We use the SUN3D dataset post-processed by \cite{ummenhofer2017demon}, which discards the samples with a high photo-consistency error.

\subsection{Depth Evaluation}
\begin{table*}[t]
\begin{centering}
\tabcolsep=0.09cm
\small
\caption{\textbf{Depth and Pose Estimation Results on MVS, Scenes11, and SUN3D Datasets}. Base-SIFT and Base-Matlab come from \cite{ummenhofer2017demon}.}
\label{tab:demon_depth_pose}
\begin{tabular}{c||ccc|cc||ccc|cc||ccc|cc}
\hline
\multirow{3}{*}{Method} & \multicolumn{5}{c||}{{\textbf{MVS Dataset} }} &\multicolumn{5}{c||}{{\textbf{Scenes11 Dataset}}} &\multicolumn{5}{c}{{\textbf{Sun3D Dataset}}} \\
 \cline{2-16}
&\multicolumn{3}{c|}{{Depth}} & \multicolumn{2}{c||}{{Pose}}&\multicolumn{3}{c|}{{Depth}} & \multicolumn{2}{c||}{{Pose}}&\multicolumn{3}{c|}{{Depth}} & \multicolumn{2}{c}{{Pose}} \\
& L1-inv &Sc-inv &L1-rel & Rot & Tran &L1-inv &Sc-inv &L1-rel & Rot & Tran&L1-inv &Sc-inv &L1-rel & Rot & Tran\\
\hline
Base-SIFT &0.056 &0.309 &0.361 &21.180& 60.516& 0.051& 0.900& 1.027 &6.179& 56.650 & 0.029& 0.290& 0.286& 7.702& 41.825\\
Base-Matlab & - &- &- &10.843& 32.736&  - &- &-& 0.917 &14.639& - &- &- &5.920& 32.298\\
COLMAP~\cite{colmap}&-&-&0.384& 7.961&23.469&-&-&0.625&4.834&10.682 &-&-&0.623&4.235&15.956  \\

DeMoN~\cite{ummenhofer2017demon} & 0.047 & 0.202 & 0.305 & 5.156 & 14.447 & 0.019 & 0.315 & 0.248 & 0.809 & 8.918 & 0.019 &0.114 &0.172& 1.801 & 18.811\\

LS-Net~\cite{clark2018ls} & 0.051 & 0.221 & 0.311 & 4.653 & 11.221 &  0.010& 0.410& 0.210 &4.653 &8.210 & 0.015 & 0.189 & 0.650& 1.521& 14.347  \\
BANet~\cite{tang2018ba} & 0.030 & 0.150 & 0.080 & 3.499 & 11.238 & 0.080 &0.210& 0.130 &3.499 &10.370 & 0.015 & 0.110 & 0.060 & 1.729 & 13.260 \\
DeepSFM~\cite{wei2019deepsfm} &  0.021 &  0.129 &  0.079 &  2.824 &  9.881  &  0.007  & 0.112 &  0.064  & 0.403 &  5.828 & 0.013 & 0.093 & 0.072 & 1.704 & 13.107 \\
Ours &\textbf{0.015}&\textbf{0.102}&\textbf{0.068}& \textbf{2.417}&\textbf{3.878}& \textbf{0.005}&\textbf{0.097} & \textbf{0.058} & \textbf{0.276}&\textbf{2.041}&\textbf{0.010}&\textbf{0.081}&\textbf{0.057}&\textbf{1.391} &\textbf{10.757} \\
\hline
\end{tabular}
\end{centering}
\end{table*}
\begin{figure*}[t]
\vspace{-2mm}
\begin{center}
   \includegraphics[width=1\linewidth]{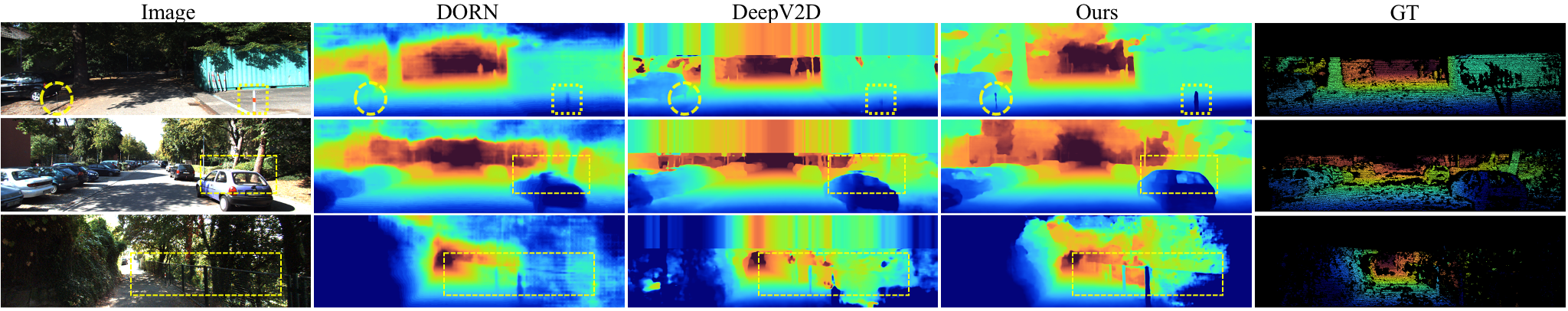}
\end{center}
\vspace{-3mm}
   \caption{\small \textbf{Qualitative Results on the KITTI Dataset.} The yellow circles and boxes in the top row highlight tiny poles which are captured more accurately by our method.  }
\label{fig:kitti_vis}
\vspace{-4mm}
\end{figure*}

We perform depth evaluation on KITTI Depth, MVS, Scenes11, and SUN3D datasets.

\textbf{KITTI Depth.} We compare our framework with both types of deep SfM methods using seven commonly used depth metrics~\cite{eigen2014depth}. We also leverage one disparity metric D1-all\footnote{Percentage of stereo disparity outliers. We convert the estimated depth to disparities using the focal length and baseline provided by KITTI.} as it measures the precision of the depth estimation. Since the Type \textbf{I} methods are self-supervised and they all perform single frame depth estimation in inference, we report the results of the state-of-the-art supervised single image depth estimation methods \cite{fu2018deep,yin2019enforcing} as they can be viewed as the  upper bounds of Type \textbf{I} methods.

Quantitative results are shown in Table~\ref{tab:kitti_depth}. Although only using a flow estimator trained on synthetic datasets, our method beats all previous methods with a clear margin on various metrics, \eg, $2.273$ versus $2.727$ in RMSE. Especially, our method largely outperforms DeepV2D although DeepV2D used ground truth camera pose and five-frame sequences for training. Note that there is a number of frames in the Eigen split that do not strictly satisfy the rigid SfM assumption such as stationary scene. When only keeping the frames that satisfy SfM assumptions, \ie, on the Eigen SfM split, our method achieves even better accuracy, with $3.1\%$ \emph{vs} $9.1\%$ in D1-all. Fig.~\ref{fig:kitti_vis} illustrates some qualitative results compared with the state-of-the-art supervised single image method~\cite{fu2018deep} and deep SfM method~\cite{teed2018deepv2d}.

\begin{figure}[t]
\begin{center}
   \includegraphics[width=1\linewidth]{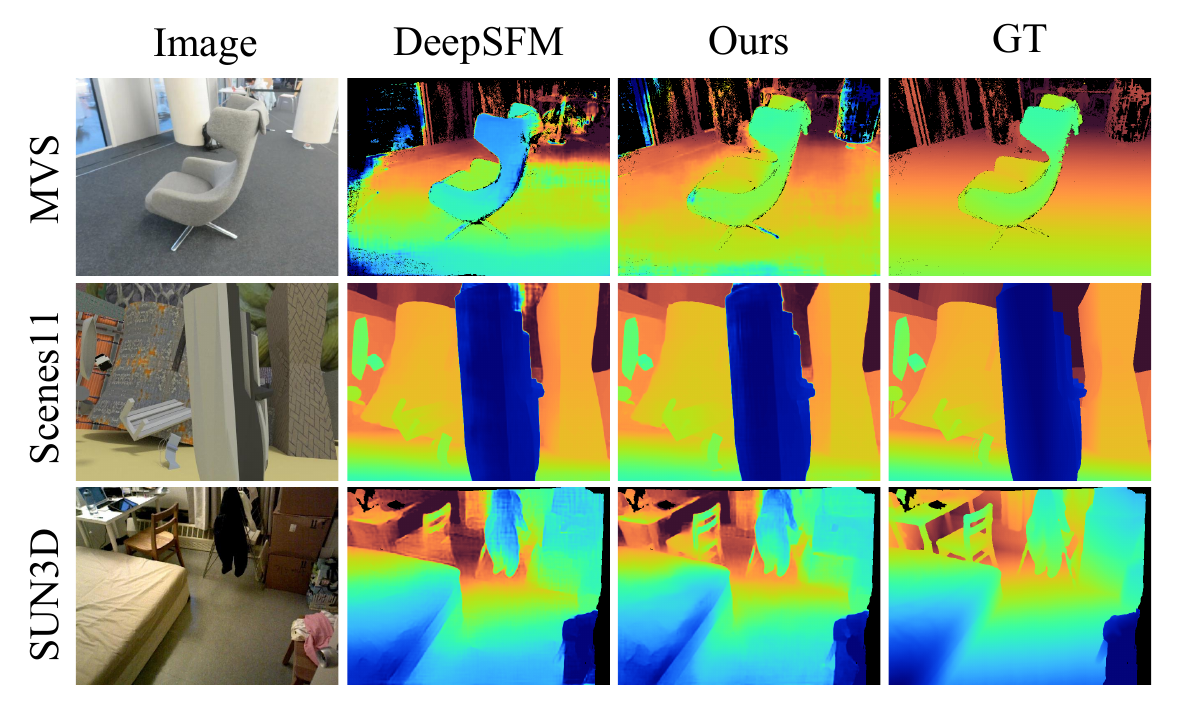}
\end{center}
\vspace{-3mm}
   \caption{\small \textbf{Qualitative Examples on MVS, Scenes11, and SUN3D Datasets,} where our method consistently achieves better results.}
\label{fig:demon_vis}
\vspace{-4mm}
\end{figure}

\textbf{MVS, Scenes11, and SUN3D.} 
We compare our framework to state-of-the-art Type \textbf{II} methods under two-view SfM setting using metrics by \cite{ummenhofer2017demon}. We use the same strategy of iterative depth refinement as \cite{wei2019deepsfm} in inference for a fair comparison. As shown in Table~\ref{tab:demon_depth_pose}, our method achieves superior performance on all metrics among all three datasets comparing with the previous state-of-the-art Type \textbf{II} methods. Fig.~\ref{fig:demon_vis} provides some qualitative results.

\subsection{Camera Pose Estimation}
We compare the camera pose estimation accuracy with Type \textbf{I} and Type \textbf{II} SfM methods on the KITTI VO, MVS, Scenes11, and SUN3D datasets.

\begin{figure}[ht]
\begin{center}
  \includegraphics[width=1\linewidth]{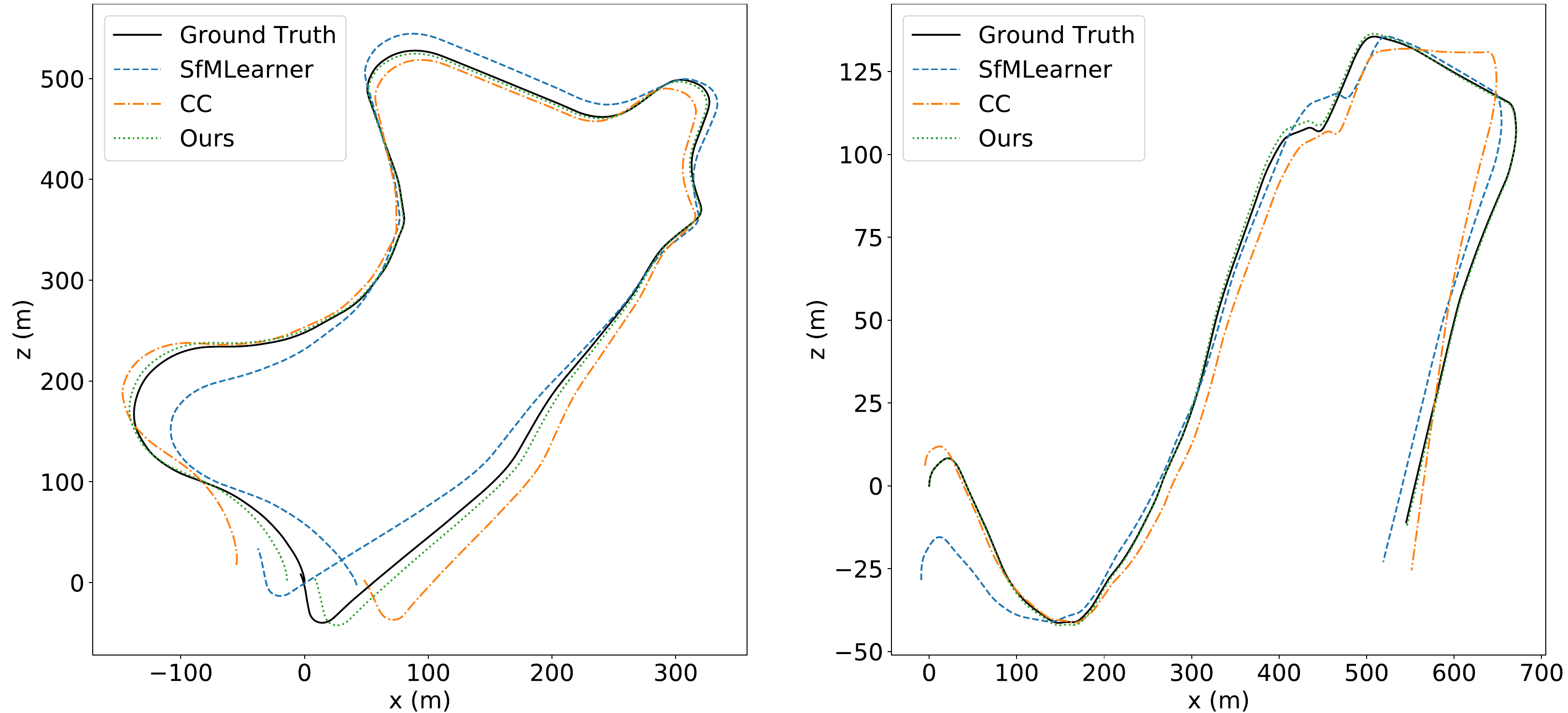}
\end{center}
\vspace{-3mm}
  \caption{\small \textbf{Visual Trajectory on the KITTI VO dataset.} We compare our method against other deep learning based SfM methods on Seq.09 (Left)  and Seq.10 (Right) of KITTI VO dataset.}
  \vspace{-3mm}
\label{fig:kitti_traj}
\end{figure}

\textbf{KITTI VO.} We measure the pose estimation accuracy on relative translational error $t_{\text{err}}$ and relative rotational error $r_{\text{err}}$ as in \cite{zou2020learning}. For all results, we align the predicted trajectories to the ground truth via least square optimization~\cite{umeyama1991least}. Our method achieves the best pose estimation accuracy with a clear margin compared with the Type \textbf{I} SfM methods \cite{sfmlearner,GANVO, Ranjan_2019_CVPR}, and full-sequence visual odometry approach~\cite{zou2020learning}. 
In Fig.~\ref{fig:kitti_traj} we visualize the full sequence odometry trajectories on $9^{\text{th}}$ and $10^{\text{th}}$ sequences. Our results are more aligned with the ground truth trajectories. It is worth noting that our model are only trained on sythetic datasets while the other methods are fine-tuned on the KITTI VO dataset and take more frames to estimate the camera poses.

\textbf{MVS, Scenes11 and SUN3D.} 
The competitors use ground truth poses to train their pose estimation module on these three datasets, while we use the ground truth poses to fine-tune our optical flow model using Eq.~\eqref{eq:flow_loss}. 
We also report the pose estimation accuracy in Table~\ref{tab:demon_depth_pose}, using the metrics of DeMoN~\cite{ummenhofer2017demon}. Our method beats the previous state-of-the-art on all three datasets with a clear margin, \eg, $60.8\%$ better in translation estimation on MVS dataset and $31.5\%$ better in rotation estimation on Scenes11 dataset. Moreover, we verify the effectiveness of rigid flow supervision, Eq~\eqref{eq:flow_loss}, in Table \ref{tab:refine-pose}. With fine-tuning, the translation errors are largely suppressed, and the rotation errors are notably reduced. It is worth noting that our model that was trained on synthetic datasets has already achieved comparable performance with previous methods.


\begin{table}
\begin{center}
\tabcolsep=0.1cm
\caption{\small\textbf{Pose Estimation Accuracy on KITTI VO dataset.} Bold indicates the best. For pose estimation, our method uses an optical flow model trained on synthetic data. The result of GANVO~\cite{GANVO} is provided by its author.}
\label{tab:kitti_pose}
\footnotesize
\begin{tabular}{c|c|c |c|c}
\hline
 \multirow{2}{*}{Method}  & \multicolumn{2}{c|}{Seq. 09} & \multicolumn{2}{c}{Seq. 10} \\
\cline{2-5}
&$t_{\text{err}}  (\%)$ &$r_{\text{err}} (^\circ/100m)$ &$t_{\text{err}}  (\%)$ &$r_{\text{err}} (^\circ/100m)$ \\
\hline
SfMLearner~\cite{sfmlearner}& 8.28 &3.07&12.20 &2.96\\
GANVO~\cite{GANVO} & 11.52 &3.53 & 11.60 & 5.17 \\
CCNet~\cite{Ranjan_2019_CVPR}&6.92 &1.77& 7.97 &3.11 \\
LTMVO~\cite{zou2020learning} &3.49&1.03&5.81&1.82 \\

Ours& \textbf{1.70} & \textbf{0.48} & \textbf{1.49} & \textbf{0.55} \\

\hline
\end{tabular}
\end{center}
\vspace{-3mm}
\end{table}

\begin{table}[ht]
\small
\tabcolsep=0.13cm
\centering
\caption{\small{\textbf{The Effect of Optical Flow Fine-tuning.} With the help of rigid flow supervision, our fine-tuned model achieves a much better camera pose result than the model trained on synthetic data. }}
\begin{tabular}{c|c|c|c|c|c|c}
\hline
\multirow{2}{*}{Model} & \multicolumn{2}{c|}{\textbf{MVS}} & \multicolumn{2}{c|}{\textbf{Scenes11}} & \multicolumn{2}{c}{\textbf{SUN3D}} \\ \cline{2-7} 
                  &   Rot        &   Tran        &   Rot        &   Tran        &   Rot        &   Tran        \\ \hline
Our-synthetic      & 3.637 & 10.984      &  0.587&6.617        &    1.670 &12.905          \\ \hline
Our-finetune              &\textbf{2.417}&\textbf{3.878}        &    \textbf{0.276}&\textbf{2.041}          &    \textbf{1.391} &\textbf{10.757}       
\\ \hline
\end{tabular}
\label{tab:refine-pose}
\vspace{-0mm}
\end{table}

\subsection{Framework Analysis and Justification}
\label{sec:ablaiton}
\textbf{Estimating Camera Pose from Optical Flow.}
There are multiple ways to extract camera pose from optical flow. We consider two kinds of methods: deep regression and the classic five-point algorithm~\cite{li2006five} with RANSAC scheme. For deep regression methods, we build a PoseNet similar to the one used in~\cite{flowmotion} with ResNet50~\cite{resnet} as the feature backbone, using image pairs and optical flow as the input. For the five-point algorithm, we use flow matching pairs as the input. We also set a baseline by using SIFT matches. To filter out error matches and outliers, we compare different masking strategies, such as flow uncertainty maps (output of per-pixel softmax operation), learned confidence maps, and SIFT feature locations.

We evaluate these methods on the MVS dataset, see Table~\ref{tab:ab_pose_from_flow}. Deep regression methods have almost constant performance regardless of different inputs and masking strategies. The best option is to use flow matches with masks based on SIFT feature locations.\footnote{Note that we use SIFT feature \emph{detection} to obtain state-of-the-art results, whereas SIFT feature \emph{matching} performs poorly and is not used.}

\textbf{Dealing With Misaligned Scales.}
It is impossible to perfectly recover absolute scales from two-view images. This scale-ambiguity problem will cause trouble if we would like to directly use ground truth depth for supervision or through the widely used scale-invariant loss~\cite{eigen2014depth,ummenhofer2017demon}. We verify the effect of the proposed scale-invariant depth estimation module on the KITTI depth Eigen split. The baseline follows our pipeline but without scale-invariant depth module. It simply uses $\ell_{huber}$ loss on the estimated depth and the ground depth regardless their scales, which forces the network to implicitly learn a scale. As shown in Table~\ref{tab:ab_depth_scale_inv_or_not}, our scale-invariant depth module achieves a very similar accuracy to `Oracle Pose', which is the upper bound of our method. On the other hand, the performance of the scale-invariant loss is similar to the baseline method, which indicates that this loss cannot handle the scale problem.


\begin{table}
\begin{center}
\tabcolsep = 0.2cm
\caption{\textbf{Estimating Camera Pose from Optical Flow.} We compare different methods to estimate camera pose from optical flow on the MVS dataset. `CNN' represents the pose regression network based on convolutional neural networks with ground truth pose supervision. `5-point' represents the five-point algorithm with RANSAC scheme. We also compare different flow masking strategies here.}
\label{tab:ab_pose_from_flow}
\footnotesize
\begin{tabular}{c | c|c|cc }
\hline
Method& Input & Sparse Mask & Rot & Tran \\
\hline
CNN & Color &-& 6.652& 17.834\\ 
CNN & Color + Flow &-& 6.437& 17.216\\
CNN & Color + Flow & Uncertainty & 6.528& 17.107 \\
CNN & Color + Flow & Confidence & 6.532& 17.511\\

CNN & Color + Flow & SIFT loc & 6.512& 17.231\\
\hline
5-point & SIFT matches & - &10.622&29.731 \\

5-point & Flow matches & - &15.673&37.292 \\
5-point & Flow matches & Uncertainty & 4.923 &12.127 \\
5-point & Flow matches & Confidence & 4.614 &11.022 \\
5-point & Flow matches & SIFT Loc &\textbf{2.417}&\textbf{3.878} \\
\bottomrule
\end{tabular}
\end{center}

\vspace{-6mm}
\end{table}

\begin{table}[htb]
\footnotesize
\tabcolsep = 0.2cm
\centering
\caption{\small{\textbf{Dealing with Misaligned Scales.} We compare different strategies to handle the misaligned scales between the estimated depth and ground truth depth on the KITTI Eigen split. `Scale Inv Matching' indicates the scale invariant matching for plane sweeping, `Scale Inv Loss' represents the scale invariant depth loss. The `Oracle' means using the ground truth for both training and inference. Using ground truth pose for training achieves a worse result than the baseline, which verifies the scaling problem. }  } 
\begin{tabular}{ c | c | c | c |c }
\hline
 Strategy & Abs Rel & Sq Rel & $\text{RMSE}$ & $\text{RMSE}_{log}$   \\
 \hline
Baseline  & 0.089&0.318& 3.120& 0.129 \\
GT Pose Training  &0.121&0.438&3.421&0.175\\
Scale Inv Loss & 0.084&0.302& 2.981& 0.116 \\
Scale Inv Matching  & 0.055&0.224&2.273 &0.091 \\
\hline
Oracle Scale  &   0.053&0.216&2.271&0.089 \\
Oracle Pose  & 0.052&0.212&2.269&0.088 \\
\hline
\end{tabular}
\label{tab:ab_depth_scale_inv_or_not}
\vspace{-2mm}
\end{table}



\textbf{Scale-invariant Matching on Other Frameworks.}
The scale-invariant matching is specifically designed for our pipeline to handle the scale ambiguity in depth estimation. Previous deep SfM methods like DeepV2D do not suffer from this problem as they force networks to regress camera poses with scales and then make the depth scaled. That means, these methods cannot benefit from the scale-invariant matching. As a proof, we apply our scale-invariant matching to DeepV2D and test it on the KITTI Eigen dataset. The performance gain is minor: Abs Rel from $0.064$ to $0.063$ and RMSE from $2.946$ to $2.938$. Our superior performance benefits from the whole newly proposed deep SfM pipeline rather than a single component. Since all components are tightly coupled in our pipeline, replacing either of them will result in a severe performance drop.

\section{Conclusion}
In this paper, we have revisited the problem of deep neural network based two-view SfM. First, we argued that existing deep learning-based SfM approaches formulate depth estimation or pose estimation as ill-posed problems. Then we proposed a new deep two-view SfM framework that follows the classic well-posed SfM pipeline. Extensive experiments show that our proposed method outperforms all state-of-the-art methods in both pose and depth estimation with a clear margin. In the future, we plan to extend our framework to other SfM problems such as three-view SfM and multi-view SfM, where the loop consistency and temporal consistency could further constrain these already well-posed problems.

\small{
\noindent
\textbf{Acknowledgements}
Yuchao Dai was supported in part by National Natural Science Foundation of China (61871325) and National Key Research and Development Program of China (2018AAA0102803). Hongdong Li was supported in part by ACRV (CE140100016), ARC-Discovery (DP 190102261), and ARC-LIEF (190100080) grants. We would like to thank Shihao Jiang, Dylan Campbell, Charles Loop for helpful discussions and Ke Chen for providing field test images from NVIDIA AV cars.}

{\small
\bibliographystyle{ieee_fullname}
\bibliography{SFM_Net}

\begin{thebibliography}{10}\itemsep=-1pt

\bibitem{agarwal2011building}
Sameer Agarwal, Yasutaka Furukawa, Noah Snavely, Ian Simon, Brian Curless,
  Steven~M Seitz, and Richard Szeliski.
\newblock Building rome in a day.
\newblock {\em Communications of the ACM}, 54(10):105--112, 2011.

\bibitem{GANVO}
Yasin Almalioglu, Muhamad Risqi~U Saputra, Pedro~PB de Gusmao, Andrew Markham,
  and Niki Trigoni.
\newblock Ganvo: Unsupervised deep monocular visual odometry and depth
  estimation with generative adversarial networks.
\newblock In {\em 2019 International conference on robotics and automation
  (ICRA)}, pages 5474--5480. IEEE, 2019.

\bibitem{bertero1988illposed}
M. Bertero, T. Poggio, and V. Torre.
\newblock Ill-posed problems in early vision.
\newblock {\em Proceedings of the IEEE}, 76(8):869--889, Aug. 1988.

\bibitem{Butler2012Sintel}
Daniel~J Butler, Jonas Wulff, Garrett~B Stanley, and Michael~J Black.
\newblock A naturalistic open source movie for optical flow evaluation.
\newblock In {\em Proceedings of the European Conference on Computer Vision
  (ECCV)}, pages 611--625. Springer, 2012.

\bibitem{chen2019self}
Yuhua Chen, Cordelia Schmid, and Cristian Sminchisescu.
\newblock Self-supervised learning with geometric constraints in monocular
  video: Connecting flow, depth, and camera.
\newblock In {\em Int. Conf. Comput. Vis.}, pages 7063--7072, 2019.

\bibitem{cheng2020hierarchical}
Xuelian Cheng, Yiran Zhong, Mehrtash Harandi, Yuchao Dai, Xiaojun Chang,
  Hongdong Li, Tom Drummond, and Zongyuan Ge.
\newblock Hierarchical neural architecture search for deep stereo matching.
\newblock {\em Advances in Neural Information Processing Systems}, 33, 2020.

\bibitem{clark2018ls}
Ronald Clark, Michael Bloesch, Jan Czarnowski, Stefan Leutenegger, and Andrew~J
  Davison.
\newblock Ls-net: Learning to solve nonlinear least squares for monocular
  stereo.
\newblock {\em arXiv preprint arXiv:1809.02966}, 2018.

\bibitem{davison2007pami:monoslam}
A.~J. {Davison}, I.~D. {Reid}, N.~D. {Molton}, and O. {Stasse}.
\newblock {MonoSLAM}: {R}eal-time single camera {SLAM}.
\newblock {\em IEEE Transactions on Pattern Analysis and Machine Intelligence
  (PAMI)}, 29(6):1052--1067, 2007.

\bibitem{flyingchair}
Alexey Dosovitskiy, Philipp Fischer, Eddy Ilg, Philip Hausser, Caner Hazirbas,
  Vladimir Golkov, Patrick Van Der~Smagt, Daniel Cremers, and Thomas Brox.
\newblock Flownet: Learning optical flow with convolutional networks.
\newblock In {\em Int. Conf. Comput. Vis.}, pages 2758--2766, 2015.

\bibitem{eigen2014depth}
David Eigen, Christian Puhrsch, and Rob Fergus.
\newblock Depth map prediction from a single image using a multi-scale deep
  network.
\newblock In {\em Advances in neural information processing systems}, pages
  2366--2374, 2014.

\bibitem{engel2016dso}
J. Engel, V. Koltun, and D. Cremers.
\newblock Direct sparse odometry.
\newblock In {\em arXiv:1607.02565}, 2016.

\bibitem{engel14eccv:lsdslam}
J. Engel, T. Schöps, and D. Cremers.
\newblock {LSD-SLAM}: Large-scale direct monocular {SLAM}.
\newblock In {\em European Conference on Computer Vision (ECCV)}, 2014.

\bibitem{fischler1981random}
Martin~A Fischler and Robert~C Bolles.
\newblock Random sample consensus: a paradigm for model fitting with
  applications to image analysis and automated cartography.
\newblock {\em Communications of the ACM}, 24(6):381--395, 1981.

\bibitem{forster2014icra:svo}
Christian Forster, Matia Pizzoli, and Davide Scaramuzza.
\newblock {SVO}: {F}ast semi-direct monocular visual odometry.
\newblock In {\em ICRA}, 2014.

\bibitem{fu2018deep}
Huan Fu, Mingming Gong, Chaohui Wang, Kayhan Batmanghelich, and Dacheng Tao.
\newblock Deep ordinal regression network for monocular depth estimation.
\newblock In {\em IEEE Conf. Comput. Vis. Pattern Recog.}, pages 2002--2011,
  2018.

\bibitem{kittidataset}
Andreas Geiger, Philip Lenz, and Raquel Urtasun.
\newblock Are we ready for autonomous driving? the kitti vision benchmark
  suite.
\newblock In {\em 2012 IEEE Conference on Computer Vision and Pattern
  Recognition}, pages 3354--3361. IEEE, 2012.

\bibitem{hartley2003multiple}
Richard Hartley and Andrew Zisserman.
\newblock {\em Multiple view geometry in computer vision}.
\newblock Cambridge university press, 2003.

\bibitem{hassner2006cvprw}
T. {Hassner} and R. {Basri}.
\newblock Example based {3D} reconstruction from single {2D} images.
\newblock In {\em Conference on Computer Vision and Pattern Recognition
  Workshop (CVPRW)}, 2006.

\bibitem{resnet}
Kaiming He, Xiangyu Zhang, Shaoqing Ren, and Jian Sun.
\newblock Deep residual learning for image recognition.
\newblock In {\em Proceedings of the IEEE conference on computer vision and
  pattern recognition}, pages 770--778, 2016.

\bibitem{horn1981determining}
Berthold~KP Horn and Brian~G Schunck.
\newblock Determining optical flow.
\newblock {\em Artificial intelligence}, 17(1-3):185--203, 1981.

\bibitem{im2019dpsnet}
Sunghoon Im, Hae-Gon Jeon, Stephen Lin, and In~So Kweon.
\newblock Dpsnet: End-to-end deep plane sweep stereo.
\newblock {\em International Conference on Learning Representations}, 2019.

\bibitem{klein2007ismar:ptam}
Georg Klein and David Murray.
\newblock Parallel tracking and mapping for small {AR} workspaces.
\newblock In {\em ISMAR}, 2007.

\bibitem{li2021arvo}
Dongxu Li, Chenchen Xu, Kaihao Zhang, Xin Yu, Yiran Zhong, Wenqi Ren, Hanna
  Suominen, and Hongdong Li.
\newblock Arvo: Learning all-range volumetric correspondence for video
  deblurring.
\newblock {\em arXiv preprint arXiv:2103.04260}, 2021.

\bibitem{li2006five}
Hongdong Li and Richard Hartley.
\newblock Five-point motion estimation made easy.
\newblock In {\em 18th International Conference on Pattern Recognition
  (ICPR'06)}, volume~1, pages 630--633. IEEE, 2006.

\bibitem{li2019bmvc:svosr}
Kejie Li, Ravi Garg, Ming Cai, and Ian Reid.
\newblock Single-view object shape reconstruction using deep shape prior and
  silhouette.
\newblock In {\em BMVC}, 2019.

\bibitem{longuet1981computer}
H~Christopher Longuet-Higgins.
\newblock A computer algorithm for reconstructing a scene from two projections.
\newblock {\em Nature}, 293(5828):133--135, 1981.

\bibitem{loweSIFT}
David~G Lowe.
\newblock Object recognition from local scale-invariant features.
\newblock In {\em Proceedings of the seventh IEEE international conference on
  computer vision}, volume~2, pages 1150--1157. Ieee, 1999.

\bibitem{Mahjourian_2018_CVPR}
Reza Mahjourian, Martin Wicke, and Anelia Angelova.
\newblock Unsupervised learning of depth and ego-motion from monocular video
  using 3d geometric constraints.
\newblock In {\em IEEE Conf. Comput. Vis. Pattern Recog.}, June 2018.

\bibitem{Mayer2016Things3D}
Nikolaus Mayer, Eddy Ilg, Philip Hausser, Philipp Fischer, Daniel Cremers,
  Alexey Dosovitskiy, and Thomas Brox.
\newblock A large dataset to train convolutional networks for disparity,
  optical flow, and scene flow estimation.
\newblock In {\em IEEE Conf. Comput. Vis. Pattern Recog.}, pages 4040--4048,
  2016.

\bibitem{mur2015orbslam}
Raul Mur-Artal, Jose Maria~Martinez Montiel, and Juan~D Tardos.
\newblock {ORB-SLAM}: {A} versatile and accurate monocular {SLAM} system.
\newblock {\em IEEE Transactions on Robotics}, 31(5):1147--1163, 2015.

\bibitem{nister2004efficient}
David Nist{\'e}r.
\newblock An efficient solution to the five-point relative pose problem.
\newblock {\em IEEE transactions on pattern analysis and machine intelligence},
  26(6):756--770, 2004.

\bibitem{Ranjan_2019_CVPR}
Anurag Ranjan, Varun Jampani, Lukas Balles, Kihwan Kim, Deqing Sun, Jonas
  Wulff, and Michael~J. Black.
\newblock Competitive collaboration: Joint unsupervised learning of depth,
  camera motion, optical flow and motion segmentation.
\newblock In {\em IEEE Conf. Comput. Vis. Pattern Recog.}, June 2019.

\bibitem{colmap}
Johannes~L Schonberger and Jan-Michael Frahm.
\newblock Structure-from-motion revisited.
\newblock In {\em Proceedings of the IEEE Conference on Computer Vision and
  Pattern Recognition}, pages 4104--4113, 2016.

\bibitem{sun2018pwc}
Deqing Sun, Xiaodong Yang, Ming-Yu Liu, and Jan Kautz.
\newblock Pwc-net: Cnns for optical flow using pyramid, warping, and cost
  volume.
\newblock In {\em Proceedings of the IEEE conference on computer vision and
  pattern recognition}, pages 8934--8943, 2018.

\bibitem{tang2018ba}
Chengzhou Tang and Ping Tan.
\newblock Ba-net: Dense bundle adjustment network.
\newblock {\em International Conference on Learning Representations}, 2018.

\bibitem{teed2018deepv2d}
Zachary Teed and Jia Deng.
\newblock Deepv2d: Video to depth with differentiable structure from motion.
\newblock {\em International Conference on Learning Representations}, 2020.

\bibitem{teed2020raft}
Zachary Teed and Jia Deng.
\newblock Raft: Recurrent all-pairs field transforms for optical flow.
\newblock {\em arXiv preprint arXiv:2003.12039}, 2020.

\bibitem{umeyama1991least}
Shinji Umeyama.
\newblock Least-squares estimation of transformation parameters between two
  point patterns.
\newblock In {\em IEEE Trans. Pattern Anal. Mach. Intell.}, pages 376--380.
  IEEE, 1991.

\bibitem{ummenhofer2017demon}
Benjamin Ummenhofer, Huizhong Zhou, Jonas Uhrig, Nikolaus Mayer, Eddy Ilg,
  Alexey Dosovitskiy, and Thomas Brox.
\newblock Demon: Depth and motion network for learning monocular stereo.
\newblock In {\em IEEE Conf. Comput. Vis. Pattern Recog.}, pages 5038--5047,
  2017.

\bibitem{zhong2020nipsflow}
Jianyuan Wang, Yiran Zhong, Yuchao Dai, Kaihao Zhang, Pan Ji, and Hongdong Li.
\newblock Displacement-invariant matching cost learning for accurate optical
  flow estimation.
\newblock In {\em Adv. Neural Inform. Process. Syst.}, 2020.

\bibitem{flowmotion}
Kaixuan Wang and Shaojie Shen.
\newblock Flow-motion and depth network for monocular stereo and beyond.
\newblock {\em IEEE Robotics and Automation Letters}, 5(2):3307--3314, 2020.

\bibitem{wei2019deepsfm}
Xingkui Wei, Yinda Zhang, Zhuwen Li, Yanwei Fu, and Xiangyang Xue.
\newblock Deepsfm: Structure from motion via deep bundle adjustment.
\newblock {\em Eur. Conf. Comput. Vis.}, 2020.

\bibitem{xiao2013sun3d}
Jianxiong Xiao, Andrew Owens, and Antonio Torralba.
\newblock Sun3d: A database of big spaces reconstructed using sfm and object
  labels.
\newblock In {\em Proceedings of the IEEE international conference on computer
  vision}, pages 1625--1632, 2013.

\bibitem{xie2019iccv:pix2vox}
Haozhe Xie, Hongxun Yao, Xiaoshuai Sun, Shangchen Zhou, and Shengping Zhang.
\newblock {Pix2Vox}: {C}ontext-aware {3D} reconstruction from single and
  multi-view images.
\newblock In {\em ICCV}, 2019.

\bibitem{yin2019enforcing}
Wei Yin, Yifan Liu, Chunhua Shen, and Youliang Yan.
\newblock Enforcing geometric constraints of virtual normal for depth
  prediction.
\newblock In {\em Int. Conf. Comput. Vis.}, pages 5684--5693, 2019.

\bibitem{yin2018geonet}
Zhichao Yin and Jianping Shi.
\newblock Geonet: Unsupervised learning of dense depth, optical flow and camera
  pose.
\newblock In {\em IEEE Conf. Comput. Vis. Pattern Recog.}, 2018.

\bibitem{zhong2017self}
Yiran Zhong, Yuchao Dai, and Hongdong Li.
\newblock Self-supervised learning for stereo matching with self-improving
  ability.
\newblock {\em arXiv preprint arXiv:1709.00930}, 2017.

\bibitem{zhong18eccvmono}
Yiran Zhong, Yuchao Dai, and Hongdong Li.
\newblock Stereo computation for a single mixture image.
\newblock In Vittorio Ferrari, Martial Hebert, Cristian Sminchisescu, and Yair
  Weiss, editors, {\em Computer Vision -- ECCV 2018}, pages 441--456, Cham,
  2018. Springer International Publishing.

\bibitem{zhong2020efficient}
Yiran Zhong, Yuchao Dai, and Hongdong Li.
\newblock Efficient depth completion using learned bases.
\newblock {\em arXiv preprint arXiv:2012.01110}, 2020.

\bibitem{zhong2019unsupervised}
Yiran Zhong, Pan Ji, Jianyuan Wang, Yuchao Dai, and Hongdong Li.
\newblock Unsupervised deep epipolar flow for stationary or dynamic scenes.
\newblock In {\em IEEE Conf. Comput. Vis. Pattern Recog.}, pages 12095--12104,
  2019.

\bibitem{Zhong_2018_ECCV}
Yiran Zhong, Hongdong Li, and Yuchao Dai.
\newblock Open-world stereo video matching with deep rnn.
\newblock In {\em Proceedings of the European Conference on Computer Vision
  (ECCV)}, September 2018.

\bibitem{zhong2020displacement}
Yiran Zhong, Charles Loop, Wonmin Byeon, Stan Birchfield, Yuchao Dai, Kaihao
  Zhang, Alexey Kamenev, Thomas Breuel, Hongdong Li, and Jan Kautz.
\newblock Displacement-invariant cost computation for efficient stereo
  matching.
\newblock {\em arXiv preprint arXiv:2012.00899}, 2020.

\bibitem{sfmlearner}
Tinghui Zhou, Matthew Brown, Noah Snavely, and David~G Lowe.
\newblock Unsupervised learning of depth and ego-motion from video.
\newblock In {\em Proceedings of the IEEE Conference on Computer Vision and
  Pattern Recognition}, pages 1851--1858, 2017.

\bibitem{zou2020learning}
Yuliang Zou, Pan Ji, , Quoc-Huy Tran, Jia-Bin Huang, and Manmohan Chandraker.
\newblock Learning monocular visual odometry via self-supervised long-term
  modeling.
\newblock In {\em European Conference on Computer Vision}, 2020.

\end{thebibliography}
}

\end{document}